%% file: main.tex
\documentclass[runningheads]{llncs}

\usepackage[T1]{fontenc}
\usepackage{amsmath}
\usepackage{amssymb}
\usepackage{booktabs}
\usepackage{float}
\usepackage{graphicx}
\usepackage{marvosym}
\usepackage{url}

\begin{document}

\title{Adaptive Skill Granularity for Coding Agents:\\
Runtime Routing and Macro-to-Micro Recovery}
\titlerunning{Adaptive Skill Granularity for Coding Agents}


\author{
Dingxuan Zhang\inst{1}\thanks{These authors contributed equally to this work. Code is available at: \protect\url{https://github.com/Jayceon-3/Adaptive-Skill-Agent}}
\and
Yanhui Jiang\inst{2\star}
\and
Jun Wang\inst{3}
\and
Hefeng Zhou\inst{1}
}



\authorrunning{Dingxuan Zhang et al.}

\institute{
Shanghai Jiao Tong University, China\\
\and
Kuaishou Technology Co., Ltd., China\\
\and
University of Warwick, UK\\
\email{h.rezin.zhou@gmail.com}
}

\maketitle

\pagestyle{empty}

\input{sections/00_frontmatter}

\input{sections/01_introduction}

\input{sections/02_method}

\input{sections/03_experiments}

\input{sections/04_related_threats_conclusion}

\subsection*{Disclosure of Interests}
\noindent The authors have no competing interests to declare that are relevant to the content of this article.

\bibliographystyle{splncs04}
\bibliography{references}

\end{document}

%% file: sections/00_frontmatter.tex
\begin{abstract}
Reusable skills allow large language model agents to package task-specific procedures, tool affordances, and execution guidance into modular units. Prior work has largely studied how agents retrieve, compose, or synthesize skills before execution. In coding agents, however, execution poses an additional runtime control problem: the same task may be solved through primitive tool calls, macro skills, or hybrid execution that starts with a macro and falls back to micro-level recovery. We introduce Adaptive Skill Granularity, a framework that treats skill granularity as an action-abstraction policy conditioned on task structure, execution state, and macro-skill failures. The system combines a feature-based granularity controller and a finite-state recovery executor. We evaluate the framework with a deterministic coding-agent protocol using verifier-defined success, total cost, and process metrics such as macro adoption, failure recovery, route match, and transition correctness. Results show that fixed macro execution is efficient when applicable but brittle under mismatch, while runtime routing and explicit macro-to-micro recovery improve robustness and recover controlled macro failures that fixed-granularity execution leaves unresolved. Our findings suggest that skill libraries for coding agents should be treated not only as retrievable resources, but also as controllable, verifiable, and recoverable execution abstractions.

\keywords{LLM agents, agent skill, adaptive skill granularity}
\end{abstract}

%% file: sections/01_introduction.tex
\section{Introduction}

Large language model (LLM) agents increasingly solve software-engineering tasks
by inspecting files, invoking tools, editing code, and validating changes. To
avoid solving each task from scratch, recent systems package reusable procedures
as skills: named artifacts that specify when to act, which tools to use, what
state to inspect, and how to verify completion. Existing skill ecosystems include
Markdown skills, structured runtime skills, skill routers, dependency-aware skill
graphs, and mined trajectory fragments~\cite{skillMining2026,formalSkill2026,skillRouter2026,skillSurvey2026}.

Most prior work treats skill use as a selection or retrieval problem: given a
task and a skill library, expose the most relevant skill text or bundle before
planning~\cite{skillWeaver2026,graphOfSkills2026,skillRouter2026}. This is
necessary but insufficient for tool-using agents. During execution, the agent
must also decide the granularity of its next actions. A simple bug fix may be
safer with micro tools such as search, editing, syntax checks, and tests, while a
module move may benefit from a macro skill that combines file movement, import
rewriting, and validation. When such a macro fails or only partially applies, the
agent needs controlled macro-to-micro recovery rather than another retrieval
step.

This motivates treating skill granularity as a runtime control variable. Recent
evidence shows that injected skills often provide uneven gains, increase cost, or
even degrade software-engineering performance when their abstraction level does
not match the task context~\cite{sweSkillsBench2026}. Hierarchical agent methods
also adapt abstraction levels, but usually at the planning level through learned
policies or repeated refinement~\cite{himac2026,adaPlanH2026}. In contrast, we
control the admissible action set at runtime: the agent may execute with micro
tools, macro skills, or a hybrid mode that downgrades to micro tools after an
observable macro failure. This keeps the controller inspectable and makes
recovery an explicit execution transition rather than an informal retry or
re-planning heuristic.

This paper studies adaptive skill granularity as a runtime control problem. An
LLM agent with a skill library must decide not only which skill description to
retrieve, but also the abstraction level of its next actions. Our central
hypothesis is that no fixed granularity is universally optimal; the best choice
depends on task structure, skill quality, execution state, and verifier
feedback. We instantiate this problem in coding agents because repositories
provide persistent state, tools expose concrete side effects, and tests offer
deterministic validation. This makes coding a high-observability testbed while
keeping the method defined over general agent trajectories, action abstractions,
and recovery transitions. We further ground macro-skill admission in a
real-world skill-library template, requiring each candidate macro to document
its scope, interface, side effects, failure behavior, verification procedure,
and fallback path before it can enter the executable library. We make three
contributions.

\begin{enumerate}
\item We formulate runtime skill granularity control as distinct from ordinary
skill retrieval. The decision is not only which skill to expose, but whether to
execute in micro, macro, or hybrid recovery mode.

\item We present an agent-control architecture, instantiated in coding agents, that combines a feature-based granularity controller with a finite-state recovery executor to manage runtime execution abstraction.

\item We define a verifier-driven GranBench-Code evaluation protocol that treats
validated success and total cost as primary metrics, and reports macro adoption,
failure recovery, route match, transition correctness, and confidence intervals.
We apply it in a pilot study of fixed-granularity, routed, and recovery-enabled
agents, while specifying the unconstrained and LLM-router baselines required for
the full study.
\end{enumerate}

\section{Related Work}
\label{sec:related-work}

\paragraph{Agent skills and routing.}
Recent work defines agent skills as reusable procedural artifacts spanning
representation, acquisition, retrieval, execution, and evolution
\cite{agentSkillsArch2026,skillSurvey2026}. Runtime-oriented systems such as
Formal Skill further package skills with metadata, schemas, executors, and
hooks~\cite{formalSkill2026}, while SkillCraft evaluates whether agents can form
and reuse higher-level tool compositions~\cite{skillCraft2026}. In parallel,
SkillRouter, Graph-of-Skills, Skill Is Not Document, and SkillWeaver study how
to retrieve, bundle, or compose relevant skills for a task
\cite{skillWeaver2026,graphOfSkills2026,skillIsNotDocument2026,skillRouter2026}.
In short, prior works primarily optimize which skills to expose. In contrast, our work controls how coarsely an agent executes at runtime—dynamically shifting between primitive tools, macro skills, and recoverable hybrid paths based on live execution states.

\paragraph{Action abstraction, monitoring, and repair.}
Our formulation is connected to classical temporal abstraction, including
options, MAXQ, and hierarchical task networks
\cite{maxq2000,htn1994,options1999}. Macro skills can be viewed as temporally
extended actions, but we do not learn an end-to-end hierarchical policy; instead,
we use an inspectable controller to constrain the admissible action set at each
run. The downgrade mechanism also relates to execution monitoring and recovery
in planning and robotics~\cite{execMonitoring2005}, and to automated program
repair after failed edits~\cite{programRepair2019}. Compared with hierarchical
LLM-agent methods such as HiMAC and self-adaptive hierarchical planning
\cite{himac2026,adaPlanH2026}, which adapt or refine plans, we treat
granularity as a verifier-grounded runtime control decision over the next
available actions.

\paragraph{Skill evolution and coding-agent evaluation.}
Trajectory-mining systems extract reusable skills from agent traces
\cite{evoSkill2026,skillMining2026,codeSkill2026,socraticSWE2026,skillComposer2026},
but mined fragments may be brittle. Our macro-evolution loop therefore requires
replay, deterministic validation, documented side effects, and fallback behavior
before a macro is admitted. For evaluation, our protocol follows the executable software-engineering tradition
of SWE-bench \cite{swebench2024} and recent calls to assess coding agents by both final correct-ness and process behavior \cite{agentBeats2026,rigorBench2026,skillAudit2026}. Furthermore, recent findings emphasize that performance 
gains from function-calling skill injection must be carefully attributed to distinguish 
genuine procedural transfer from mere interface compliance \cite{chen2026attributing}. 
Closest in motivation, SWE-Skills-Bench reports that skill injection can yield uneven 
gains and higher cost depending on abstraction level and context fit \cite{sweSkillsBench2026}. 
Adaptive Skill Granularity targets this failure mode by controlling when skills should 
be used, avoided, or downgraded during execution.


%% file: sections/02_method.tex
\section{Problem Formulation}
\label{sec:problem-formulation}

\subsection{Agentic Execution Substrate}

We model a coding-agent episode as closed-loop execution over a repository. A
task is $\tau=(p,W_0,M,V)$, with prompt $p$, initial workspace $W_0$, metadata
$M$, and deterministic validator $V$. A run produces a trajectory
$\pi=\langle a_1,\ldots,a_T\rangle$, where each action invokes a skill from
$\mathcal{S}=\mathcal{S}_{\mu}\cup\mathcal{S}_{M}$. Primitive skills
$\mathcal{S}_{\mu}$ include file inspection, search, editing, syntax checks, and
tests; macro skills $\mathcal{S}_{M}$ bundle multiple primitive operations into
a temporally extended action. Each action transforms the workspace and yields an
observation,
\begin{equation}
\label{eq:workspace-transition}
  W_t = T_{a_t}(W_{t-1}), \qquad
  o_t = O(W_{t-1}, a_t, W_t, V),
\end{equation}
where $o_t$ records workspace changes, validation failures, and macro-specific
failure signals. These observations expose the runtime state that must be
controlled after skill retrieval.

\subsection{Granularity as Action-Abstraction Control}

A granularity mode constrains the admissible action set at a given point in an
agent trajectory. We consider $\mathcal{G}=\{\mu,M,H\}$, where $\mu$ denotes
micro execution, $M$ denotes macro execution, and $H$ denotes hybrid execution.
Given the history $h_t=(\tau,W_0,a_{1:t-1},o_{1:t-1})$, the admissible action
envelope is
\begin{equation}
\label{eq:admissible-actions}
  \mathcal{A}(g,h_t)=
  \begin{cases}
    \mathcal{S}_{\mu}, & g=\mu,\\
    \mathcal{S}_{M}, & g=M,\\
    \mathcal{S}_{M}, & g=H \land \neg F(h_t),\\
    \mathcal{S}_{\mu}, & g=H \land F(h_t),
  \end{cases}
\end{equation}
where $F(h_t)$ indicates that the history contains an observable recoverable
macro failure. Eq.~\eqref{eq:admissible-actions} is the main formal distinction
between our setting and ordinary skill retrieval: the controller changes the
action abstraction available to the downstream agent.

Operationally, the three modes have the following interpretation:

\begin{itemize}
\item \textbf{Micro mode} ($g=\mu$): admissible actions are restricted to
$\mathcal{S}_{\mu}$.
\item \textbf{Macro mode} ($g=M$): admissible actions are restricted to
$\mathcal{S}_{M}$.
\item \textbf{Hybrid mode} ($g=H$): execution may start with macro actions but
can transition to primitive actions after an observable failure or validation
signal.
\end{itemize}

The instantiation studied in this paper realizes hybrid mode as a one-way
macro-to-micro downgrade, which captures the dominant failure pattern we observe:
a coarse macro fails and the agent must recover with reliable primitive tools.
Eq.~\eqref{eq:admissible-actions} is, however, a special case of a more general
mode-switching policy that could also escalate from micro to macro or alternate
between modes; we treat richer switching as a direct extension of the same
admissible-action formalism rather than a separate mechanism.

A pre-execution controller selects an initial strategy
\begin{equation}
\label{eq:granularity-controller}
  d_0 = \rho(\phi(\tau)) =
  (g_0, r_0, q_0),
\end{equation}
where $\phi(\tau)$ is a feature vector extracted from the prompt, metadata, and
initial workspace; $g_0 \in \{\mu, M, H\}$ is the selected strategy; $r_0$
indicates whether macro-failure recovery is enabled; and $q_0$ is an
explanation or confidence record used for audit. The LLM agent then samples or
selects the next action only from the active envelope:
\begin{equation}
\label{eq:agent-policy}
  a_t \sim \pi_{\psi}(\cdot \mid h_t,g_t),
  \qquad a_t \in \mathcal{A}(g_t,h_t).
\end{equation}
During execution, a recovery transition function updates the effective mode:
\begin{equation}
\label{eq:recovery-transition}
  g_{t+1} = \Delta(g_t,o_t,r_0)=
  \begin{cases}
    \mu, & g_t \in \{M,H\} \land r_0=1 \land F(o_t),\\
    \bot, & g_t \in \{M,H\} \land r_0=0 \land F(o_t),\\
    g_t, & \text{otherwise}.
  \end{cases}
\end{equation}
Here $F(o_t)$ is the macro-failure predicate induced by the current observation
and $\bot$ denotes terminal macro failure.
Eq.~\eqref{eq:recovery-transition} makes recovery an explicit state transition
rather than an informal retry heuristic.

\subsection{Retrieval, Granularity, and Objective}

Skill retrieval selects a subset $\mathcal{R}_{\tau}\subseteq\mathcal{S}$ for a
task, but it does not decide whether execution should use primitive actions,
macro actions, or a recoverable mixture of both. A relevant macro may be too
coarse for the current workspace, while primitive actions may be reliable but
costly for standardized refactors. Thus, granularity control is a runtime policy
over admissible actions and recovery transitions, not just a ranking over skill
descriptions.

We define task success by the verifier,
$\mathrm{Succ}(\pi,\tau)=\mathbb{I}[V(W_T)=1]$, and view granularity control as
optimizing success under cost and unrecovered-failure penalties:
\begin{equation}
\label{eq:utility-objective}
  J(\rho,\pi_{\psi}) =
  \mathbb{E}_{\tau \sim \mathcal{D}}
  \left[
    \mathrm{Succ}(\pi,\tau)
    - \lambda_T T
    - \lambda_C \widehat{C}(\pi)
    - \lambda_F U(\pi,\tau)
  \right],
\end{equation}
where $\widehat{C}(\pi)$ is normalized token cost and $U(\pi,\tau)$ marks an
unrecovered macro failure. We report these terms through validated success,
trajectory length, token cost, macro adoption, recovery rate, route match, and
transition correctness. As a secondary summary, we also use
\begin{equation}
\label{eq:gscore-theory}
  \mathrm{G\text{-}score} =
  \frac{\mathbb{E}[\mathrm{Succ}]}
  {\mathbb{E}[T \mid \mathrm{Succ}=1] \cdot C_{\mathrm{norm}}},
\end{equation}
with token cost normalized to the micro-only baseline. G-score is used only to
compare cost among policies with similar success.

\section{Adaptive Skill Granularity}
\label{sec:method}

\begin{figure}[H]
\centering
\includegraphics[width=0.95\linewidth]{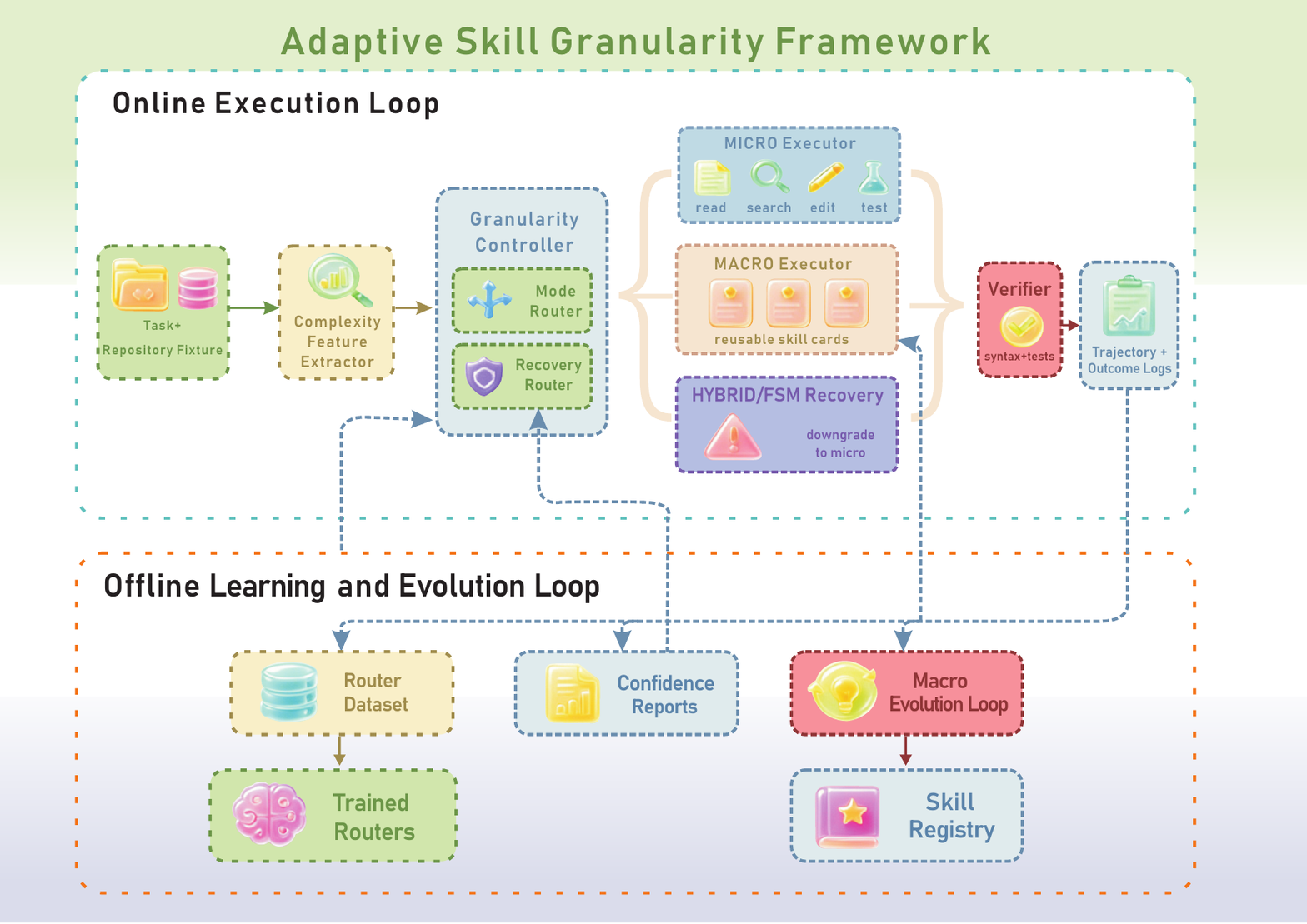}
\caption{Adaptive Skill Granularity framework. Online execution selects and
applies a micro, macro, or hybrid granularity mode, then validates the final
workspace. Offline logs support router updates, confidence reporting, and
macro-skill registry decisions.}
\label{fig:framework}
\end{figure}

Figure~\ref{fig:framework} summarizes the Adaptive Skill Granularity framework.
The online loop extracts pre-execution task features, applies the granularity
controller, restricts the downstream agent to the selected action granularity,
and validates the final workspace with deterministic checks. Execution logs are
then used for evaluation, router updates, and future macro-skill mining.

\subsection{Design Invariants}

The framework is constrained by three invariants that make the evaluation
auditable.

\begin{enumerate}
\item \textbf{Pre-execution routing only uses pre-execution evidence.} The
controller in Eq.~\eqref{eq:granularity-controller} may inspect task metadata
and the initial fixture, but not verifier outcomes from the current run.
This information separation is conceptually analogous to prior-aware evaluation in RAG-based fact-checking~\cite{sun2026pave}, which distinguishes pre-evidence model states from post-retrieval evidence.”

\item \textbf{Recovery is event-driven.} The system may transition from macro
to micro mode only after an observable macro failure, validation failure,
fault-injection event, or explicit recovery-router decision, matching
Eq.~\eqref{eq:recovery-transition}.

\item \textbf{Success is verifier-defined.} A run is successful only when the
deterministic validator passes on the final workspace, regardless of whether
the model declares completion.
\end{enumerate}

\subsection{Algorithmic Contract}

For each task, the controller first extracts complexity features and produces
a decision tuple
\begin{equation}
\label{eq:runtime-decision}
  d = (\mathrm{selected\_mode}, \mathrm{selected\_strategy},
  \mathrm{enable\_macro\_recovery}, \mathrm{reason}).
\end{equation}

The executor initializes the tool set from $d$. In macro mode, if a failure occurs and recovery is enabled, the FSM switches to micro tools and continues from the current workspace; otherwise, the failure is terminal. Every transition is appended to the trajectory log before the next action. Eq.~\eqref{eq:runtime-decision} records this runtime execution of Eq.~\eqref{eq:granularity-controller}, explicitly separating learned routing, recovery decisions, and validation evidence for analysis.

\subsection{Complexity Features}

The current implementation uses low-dimensional structural features extracted
before execution:
\begin{equation}
\label{eq:complexity-features}
  \phi(\tau) =
  [n_{\mathrm{path}}, n_{\mathrm{py}}, n_{\mathrm{nonpy}},
  n_{\mathrm{entity}}, n_{\mathrm{AST}}, d_{\mathrm{AST}},
  n_{\mathrm{import}}, n_{\mathrm{dep}}, b_{\mathrm{text}},
  b_{\mathrm{refactor}}].
\end{equation}
These dimensions capture number of source paths, Python and non-Python file
counts, entity-like strings in the prompt, AST node count, maximum AST depth,
import count, cross-file dependency count, text-task flag, and refactor-task
flag.
These features intentionally avoid execution-derived information so that the
router does not leak validation outcomes into the decision.

\subsection{Granularity Controller}

The controller supports heuristic and learned routing. The heuristic is a hand-written rule set used as a reference baseline.
The learned router, trained on prior execution logs, predicts whether a task starts in micro or macro mode, while an optional recovery router decides if a macro-compatible task should enable hybrid fallback. Together, these routers determine $g_0$ and $r_0$ in Eq.~\eqref{eq:granularity-controller}, while $q_0$ logs confidence or explanations for auditing.

To ensure interpretability and prevent memorization, the learned router is a shallow Decision Tree Classifier with a maximum depth of 4. It is trained offline on empirical success and token cost data, then serialized into JSON. At runtime, the controller deterministically traverses this lightweight JSON tree using runtime features, eliminating online retraining and opaque neural checkpoints while guaranteeing full auditability.

\subsection{Macro-Skill Interface}

We treat a macro skill as an option-like, temporally extended tool call with an
explicit contract:
\begin{equation}
\label{eq:macro-contract}
  m =
  (\mathrm{name}, \mathrm{args}, \mathrm{pre}, \mathrm{effect},
  V_m, F_m, \mathrm{fallback}),
\end{equation}
where $\mathrm{pre}$ describes when the macro is applicable,
$\mathrm{effect}$ describes allowed workspace side effects, $V_m$ is an internal
verification routine, $F_m$ is the set of observable failure classes, and
$\mathrm{fallback}$ records the intended micro-level continuation. This contract
is what lets a macro participate in the recovery transition of
Eq.~\eqref{eq:recovery-transition}: a failed macro must expose enough state for
the executor to decide whether the current workspace can be repaired with
primitive tools.

The current prototype uses three hand-written macros. \texttt{inspect\_python\_module}
summarizes a module's symbols, syntax health, and optional pattern matches;
\texttt{replace\_and\_verify} applies a targeted text replacement and then runs
syntax checks and an optional task-specific test; and
\texttt{move\_module\_and\_update\_imports} moves a Python module, rewrites
import sites, and validates the resulting workspace. These macros are
deliberately small: they are large enough to expose granularity trade-offs, but
constrained enough that their preconditions, side effects, and verifier behavior
can be audited.

\subsection{FSM Recovery Executor}

The recovery executor treats macro failure as a state transition rather than a
terminal event. Macro calls can fail because the expected text is absent, syntax
checks fail, tests fail, or a fault-injection rule simulates a brittle skill.
When recovery is enabled, the executor downgrades to primitive tools and asks
the model to continue from the current workspace. Each run records transition
count, macro failure count, recovery trigger, first failure point, and final
validated success. This executor is the concrete implementation of
Eq.~\eqref{eq:recovery-transition}.

\begin{figure}[H]
\centering
\includegraphics[width=0.88\linewidth]{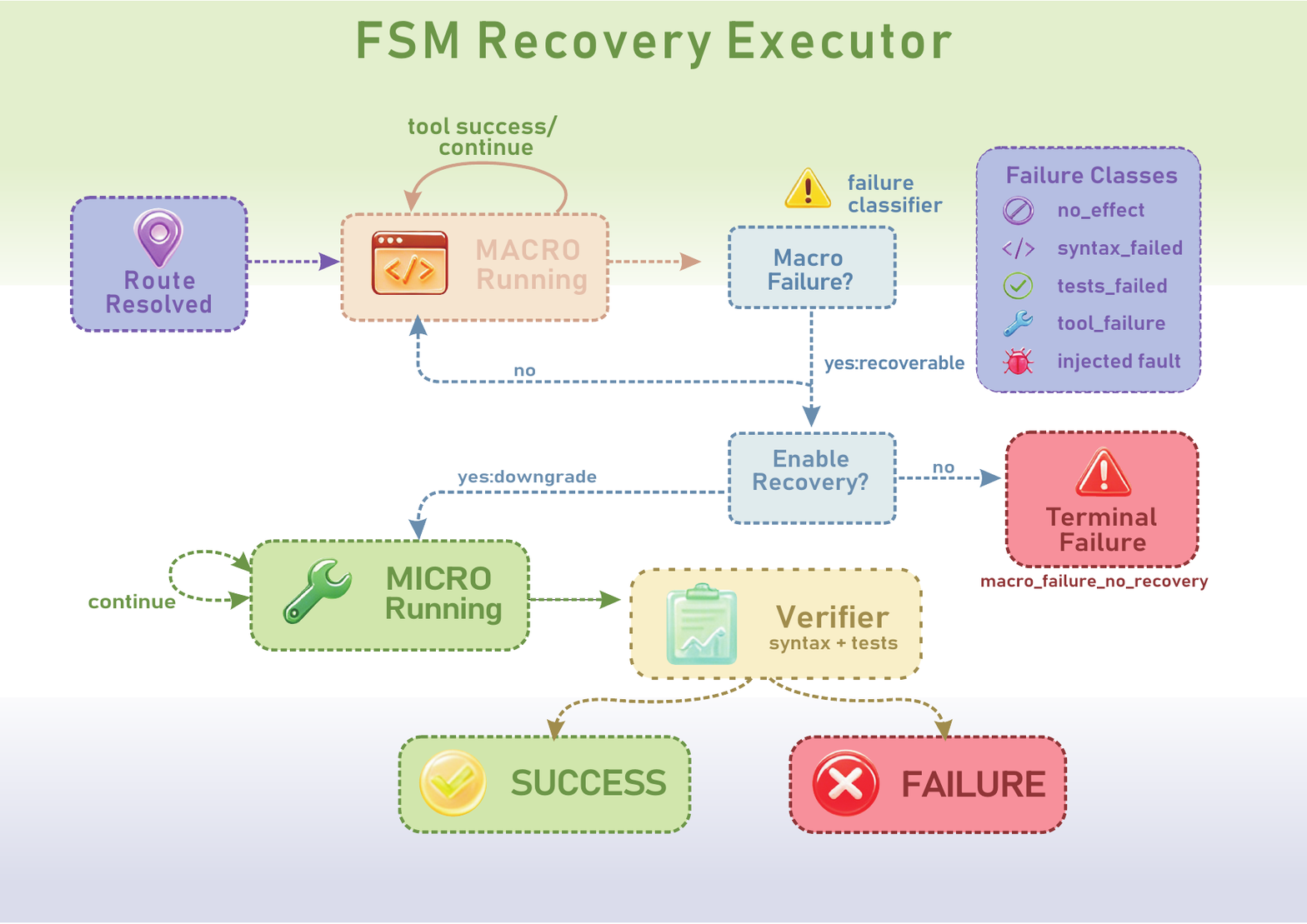}
\caption{Finite-state recovery executor. Macro runs downgrade to micro tools
after recoverable failures when recovery is enabled; otherwise they terminate
with \texttt{macro\_failure\_no\_recovery}. Success requires verifier
acceptance.}
\label{fig:fsm-recovery}
\end{figure}

\paragraph{Failure taxonomy.}
For analysis, each macro failure should be assigned one of five labels:
\emph{precondition mismatch} when the macro does not apply to the current
workspace, \emph{partial side effect} when the macro changes state before
failing, \emph{validation failure} when syntax or tests reject the result,
\emph{tool-schema failure} when arguments or returned JSON are malformed, and
\emph{executor exception} when the runtime itself fails. Recovery should be
credited only when the downgraded micro trajectory produces a validated final
state.

\subsection{Macro Evolution Loop}

While this paper focuses on the runtime routing and recovery of existing skills, a practical coding agent must eventually acquire new macro skills. We conceptualize a macro evolution loop as a natural extension of the Adaptive Skill Granularity framework, leaving its end-to-end empirical evaluation to future work.

To ensure that newly mined skills remain compatible with our FSM recovery executor, simple trajectory mining is insufficient. We propose that any generated macro candidate must pass a strict, verifier-grounded admission gate. Specifically, a candidate must: successfully replay on historical tasks, explicitly document allowable workspace side effects, and define a precise micro-level fallback path or explicitly declare an unrecoverable failure class. By enforcing this conceptual admission gate, future automated skill discovery systems can prevent brittle prompt fragments from entering the library, preserving the controllable and verifiable execution invariants required by our framework.

%% file: sections/03_experiments.tex
\section{Experiments}
\label{sec:evaluation-setting}

\subsection{Benchmark Design}

The target benchmark, GranBench-Code, categorizes tasks into four families: simple, medium, refactor, and recovery (detailed in Table~\ref{tab:families}). Each task provides an isolated workspace, source files in a failing or stale state, task metadata, and a deterministic validator. The recovery family specifically incorporates controlled failure injection before or after macro execution.

\begin{table}[t]
\centering
\scriptsize
\caption{GranBench-Code task families}
\label{tab:families}
\begin{tabular}{lcp{9cm}}
\toprule
Family & Target count & Purpose \\
\midrule
Simple & 20 & Single-file bugs, documentation examples, and configuration fixes. \\
Medium & 20 & Multi-file changes with moderate dependency structure. \\
Refactor & 20 & Module moves, symbol renames, import rewrites, and API migration. \\
Recovery & 20 & Macro failure, partial execution contamination, and downgrade recovery. \\
\bottomrule
\end{tabular}
\end{table}

\subsection{Execution Harness}

We evaluate all variants within a unified execution harness. For each task, the agent interacts with a disposable \texttt{after} workspace, while a pristine \texttt{before} snapshot is preserved for auditing. The prompt provides the task description, workspace context, active tools, and routing decisions. The agent outputs a single JSON-formatted action (specifying the skill, arguments, and reasoning) per step, or a termination signal.

The primitive tool set $\mathcal{S}_{\mu}$ includes standard file-system inspection, editing operations, and test execution. The macro tool set $\mathcal{S}_{M}$ comprises high-level composite skills, such as module inspection, verified replacement, and import-aware refactoring. All variants share the same underlying tool implementations and differ only in their exposed action spaces. Task success is determined by running task-specific validators on the final workspace. We track detailed run logs, including tool usage, routing decisions, validation results, and total token consumption across all steps.

\subsection{Baselines}

We compare several variants using identical LLMs, tools, and verifiers:
\begin{itemize}
    \item \textbf{Fixed granularity:} \texttt{A-Micro} uses only primitive skills $\mathcal{S}_{\mu}$, and \texttt{B-Macro} uses only macro skills $\mathcal{S}_{M}$. Neither variant performs recovery.
    \item \textbf{Pre-execution routing (No Recovery):} \texttt{C-Heuristic} selects the initial mode using a heuristic rule over pre-execution features. \texttt{D-Learned-ModeOnly} uses a decision-tree router trained on empirical prior logs.
    \item \textbf{Recovery-enabled hybrid execution:} \texttt{E-FSM-Downgrade} uses heuristic routing but explicitly enables a macro-to-micro downgrade transition upon macro failure. \texttt{F-Learned-Full} employs both the learned mode router and a learned recovery router.
\end{itemize}

\subsection{Metrics}

The primary evaluation metric is the validated success rate. To assess efficiency, we compute the cost required per solved task by aggregating the token cost across all runs (including failures):
\begin{equation}
\label{eq:cost-per-success}
  \mathrm{CostPerSuccess} =
  \frac{\sum_{i} C_i}{\sum_{i} \mathrm{Succ}(\pi_i,\tau_i)},
\end{equation}
where $C_i$ is the token cost of run $i$, and $\mathrm{Succ}(\pi_i,\tau_i)$ is the binary success indicator. 

To evaluate the success-cost trade-off, we also report the empirical G-score:
\begin{equation}
\label{eq:gscore-empirical}
  \mathrm{G\text{-}score} =
  \frac{\mathrm{success\_rate}}
  {\mathrm{mean\_successful\_steps} \times \mathrm{token\_cost\_normalized}}.
\end{equation}

Alongside success and cost metrics, we report process metrics including mean trajectory length, macro adoption rate, failure recovery rate, routing alignment with the rule, and transition correctness. These metrics are presented jointly to provide a comprehensive view of policy performance, efficiency, and execution behavior.

\subsection{Fixed vs. Adaptive Skill Granularity}

We first examine the trade-off between robustness and efficiency across different execution granularities (Table~\ref{tab:q1_fixed}). Pure micro execution achieves perfect success but incurs the highest token and step costs. Conversely, pure macro execution significantly reduces costs but struggles with robustness, succeeding on fewer tasks. While heuristic routing attempts to balance these extremes, it misclassifies certain tasks based on structural features. This highlights the necessity of adaptive skill granularity over relying on a fixed execution mode.

\begin{table}[ht]
\centering
\caption{Fixed-granularity and heuristic routing results (12 mixed tasks). Pure micro is robust but costly, while pure macro is efficient but brittle.}
\label{tab:q1_fixed}
\begin{tabular}{lrrrrr}
\toprule
Variant & Success & Mean Steps & Mean Tokens & Macro Adoption & G-score \\
\midrule
A-Micro  & 1.000 & 6.00 & 7217.75 & 0.000 & 0.1667 \\
B-Macro  & 0.583 & 2.71 & 3609.00 & 0.916 & 0.4305 \\
C-Heuristic & 0.750 & 3.22 & 4097.67 & 0.833 & 0.4103 \\
\bottomrule
\end{tabular}
\end{table}

\subsection{Dynamic Recovery after Macro Failure}

Next, we investigate the need for dynamic recovery upon macro failure by injecting controlled faults into the recovery benchmark (Table~\ref{tab:q2_recovery}). For early failures (\textit{Before}), lacking a recovery mechanism completely halts progress. Enabling an explicit FSM downgrade successfully recovers the tasks by falling back to micro tools, albeit at a higher token cost. 

For late failures (\textit{After}), the macro successfully updates the workspace before the failure is raised. Consequently, the no-recovery baseline passes validation without requiring additional actions. In this scenario, the unconditional FSM downgrade blindly reverts to micro mode, unnecessarily inflating token costs and execution steps. This tension demonstrates that while unconditional downgrades guarantee robustness, their overhead necessitates a more selective recovery policy.

\begin{table}[ht]
\centering
\caption{Dynamic downgrade under injected macro failures (7 tasks). The FSM downgrade recovers early failures perfectly but wastes tokens on late failures.}
\label{tab:q2_recovery}
\begin{tabular}{llrrrr}
\toprule
Fault Mode & Variant & Success & Mean Steps & Mean Tokens & Transition Rate \\
\midrule
Before & C-Heuristic-NoRecovery & 0.000 & - & -        & 0.000 \\
Before & E-FSM-Downgrade     & 1.000 & 6.86 & 11293.43 & 1.000 \\
\midrule
After  & C-Heuristic-NoRecovery & 1.000 & 2.71 & 2440.14  & 0.000 \\
After  & E-FSM-Downgrade     & 1.000 & 5.71 & 7566.71  & 1.000 \\
\bottomrule
\end{tabular}
\end{table}

\subsection{Learning Better Mode Routing}

To address the limitations of heuristic routing, we introduce a learned mode router trained offline on historical execution logs. Table~\ref{tab:q3_mode} shows that learning from empirical outcomes allows the router to consistently select the optimal execution mode, correcting misclassifications made by the heuristic policy. For instance, the learned router correctly assigns complex synchronization tasks to primitive execution, bypassing the structural features that mislead the heuristic router. Consequently, the learned router achieves full success on the evaluated tasks, demonstrating that routing decisions are better derived from empirical execution outcomes than from static rules.

\begin{table}[ht]
\centering
\caption{Learned vs. Heuristic mode routing (12 tasks). Aligning with empirical labels allows the learned router to achieve 100\% success where the heuristic fails.}
\label{tab:q3_mode}
\begin{tabular}{lrrrrr}
\toprule
Variant & Success & Mean Steps & Mean Tokens & Heuristic Match & Emp. Match \\
\midrule
C-Heuristic-ModeOnly   & 0.750 & 3.11 & 4172.89 & 1.000 & 0.667 \\
D-Learned-ModeOnly  & 1.000 & 4.17 & 5518.33 & 0.667 & 1.000 \\
\bottomrule
\end{tabular}
\end{table}

\subsection{End-to-End Adaptive Skill Granularity}

Finally, we evaluate the full Adaptive Skill Granularity framework by combining the learned mode router with the learned recovery policy. Table~\ref{tab:q4_full} shows that this combination preserves recovery performance while reducing unnecessary execution cost. Under \textit{Before} faults, F-Learned-Full matches the perfect success rate of E-FSM-Downgrade with lower token cost. Under \textit{After} faults, the recovery router skips fallback when the macro has already completed the required changes, reducing both the transition rate and token consumption compared with unconditional FSM downgrade.

The same behavior is useful for high-risk structural changes such as module moves and import rewrites. For \texttt{module\_move\_refactor}, the refactor-task feature routes execution through hybrid mode, allowing the agent to recover when strict no-recovery execution fails. These results show that learned mode selection and selective recovery complement each other: the former avoids unsuitable macro execution, while the latter limits fallback to cases where it is actually needed.

\begin{table}[ht]
\centering
\caption{Full end-to-end routing under fault injection (7 tasks). The learned full controller preserves perfect recovery success while significantly reducing token cost.}
\label{tab:q4_full}
\begin{tabular}{llrrrr}
\toprule
Fault Mode & Variant & Success & Mean Tokens & G-score & Transition Rate \\
\midrule
Before & C-Heuristic-NoRecovery & 0.000 & -        & -      & 0.000 \\
Before & E-FSM-Downgrade     & 1.000 & 11293.43 & 0.0845 & 1.000 \\
Before & F-Learned-Full      & 1.000 & 9763.57  & 0.1061 & 0.857 \\
\midrule
After  & C-Heuristic-NoRecovery & 1.000 & 2280.00  & 1.1866 & 0.000 \\
After  & E-FSM-Downgrade     & 1.000 & 7566.71  & 0.2142 & 1.000 \\
After  & F-Learned-Full      & 1.000 & 3694.43  & 0.5720 & 0.143 \\
\bottomrule
\end{tabular}
\end{table}

Overall, the proposed framework successfully combines adaptive mode selection with selective recovery, achieving consistently high task success while significantly reducing execution cost under realistic failure scenarios.

\subsection{Results Analysis}
\label{sec:current-pilot-results}

\begin{figure}[t]
  \centering
  \resizebox{0.85\linewidth}{!}{%
    \includegraphics{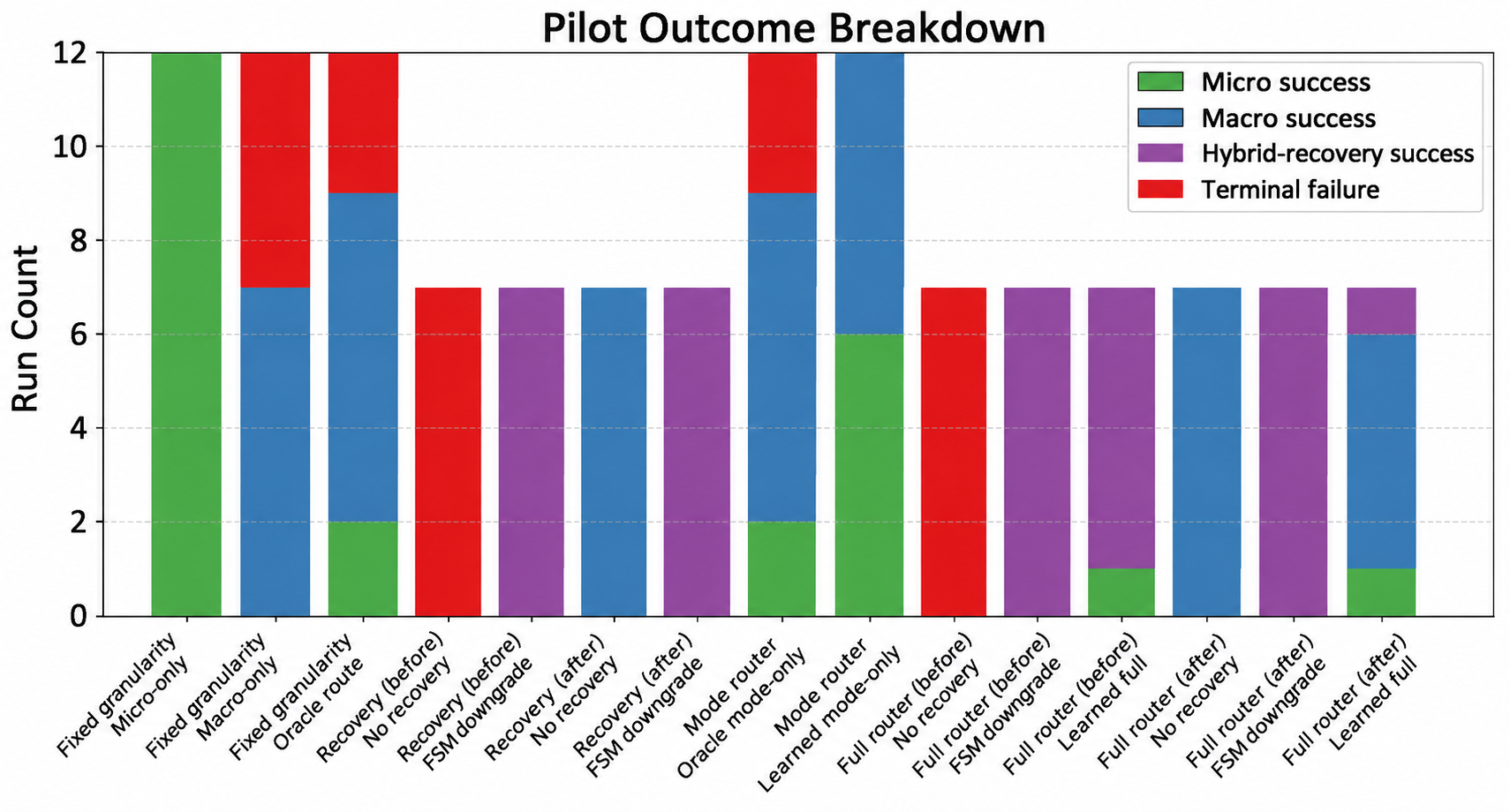}%
  }
  \caption{Evaluation outcome breakdown. Bars separate micro success, macro success, hybrid-recovery success, and terminal failure.}
  \label{fig:pilot-outcome-breakdown}
\end{figure}

On the mixed benchmark, fixed granularity exposes a clear robustness--efficiency trade-off: \texttt{A-Micro} achieves perfect success but incurs the highest step and token costs, whereas \texttt{B-Macro} is more efficient but substantially less robust, with \texttt{C-Heuristic} providing only a partial compromise. The recovery experiments further show that the value of dynamic downgrade depends on failure timing. Under early fault injection, \texttt{C-Heuristic-NoRecovery} fails completely, while \texttt{E-FSM-Downgrade} recovers all tasks at the cost of additional execution overhead. Under late faults, however, macros often complete the required edits before the failure is raised, allowing \texttt{NoRecovery} to succeed at much lower cost and making unconditional downgrade unnecessarily expensive. Together, these results indicate that neither fixed granularity nor always-on recovery is uniformly effective, motivating adaptive routing and selective recovery based on task structure and runtime failure signals.

The learned mode router demonstrates that empirical outcome labels provide a stronger basis for routing than hand-written heuristics. \texttt{D-Learned-ModeOnly} solves all evaluated tasks by correctly selecting micro execution for risky cases where the heuristic policy incorrectly relies on macros. Building on this improvement, the full learned controller (\texttt{F-Learned-Full}) further balances robustness and efficiency through selective recovery. As shown in Figure~\ref{fig:pilot-outcome-breakdown}, it matches the perfect success of \texttt{E-FSM-Downgrade} under early faults while reducing execution cost, and avoids unnecessary recovery under late faults while preserving full success and substantially lowering token consumption.

Overall, the evaluation reveals a consistent pattern: fixed granularity exposes a robustness--efficiency trade-off, while adaptive routing and recovery mitigate its two main failure modes. Dynamic downgrade is essential when macro failures occur before useful progress is made, whereas learned routing and selective recovery reduce unnecessary macro use and over-recovery when finer-grained execution is more appropriate. These results suggest that combining empirical mode selection with failure-aware recovery provides a more robust and cost-efficient execution policy than either fixed granularity or unconditional fallback alone.

%% file: sections/04_related_threats_conclusion.tex
\section{Conclusion}
\label{sec:conclusion}

Skill libraries make LLM agents more reusable, but retrieval alone does not resolve how an agent should execute a task once a relevant skill has been selected. Adaptive Skill Granularity addresses this runtime choice by allowing coding agents to operate with micro tools, macro skills, or macro execution followed by micro-level recovery. Our results show a clear trade-off: micro execution is robust but costly, while macro execution is more efficient but more sensitive to task mismatch and failure timing. Macro-to-micro recovery improves robustness under recoverable failures, and learned routing reduces unnecessary macro use and over-recovery. These results suggest that adapting execution granularity at runtime can provide a better balance between reliability and efficiency than relying on a fixed execution mode. Future work will extend GranBench-Code across broader task sets, repositories, failure modes, and models.